\pdfoutput=1
\documentclass{bmvc2k}
\usepackage{amsmath}
\usepackage{amssymb}

\makeatletter
\def\thickhline{%
  \noalign{\ifnum0=`}\fi\hrule \@height \thickarrayrulewidth \futurelet
   \reserved@a\@xthickhline}
\def\@xthickhline{\ifx\reserved@a\thickhline
               \vskip\doublerulesep
               \vskip-\thickarrayrulewidth
             \fi
      \ifnum0=`{\fi}}
\makeatother

\newlength{\thickarrayrulewidth}
\title{Learning from Counting: Leveraging Temporal Classification for Weakly Supervised Object Localization and Detection}

\addauthor{Chia-Yu Hsu}{chsu53@asu.edu}{1}
\addauthor{Wenwen Li*}{wenwen@asu.edu}{1}

\addinstitution{School of Geographical Science and Urban Planning, Arizona State University\\ Tempe, AZ 85281 USA}

\runninghead{Hsu, Li}{Leveraging temporal classification for WSOD}

\begin{document}

\maketitle

\begin{abstract}
This paper reports a new solution of leveraging temporal classification to support weakly supervised object detection (WSOD). Specifically, we introduce raster scan-order techniques to serialize 2D images into 1D sequence data, and then leverage a combined LSTM (Long, Short-Term Memory) and CTC (Connectionist Temporal Classification) network to achieve object localization based on a total count (of interested objects). We term our proposed network LSTM-CCTC (Count-based CTC). This “learning from counting” strategy differs from existing WSOD methods in that our approach automatically identifies critical points on or near a target object. This strategy significantly reduces the need of generating a large number of candidate proposals for object localization. Experiments show that our method yields state-of-the-art performance based on an evaluation on PASCAL VOC datasets.  
\end{abstract}

\section{Introduction}
\label{sec:intro}

Object detection (OD) using deep learning, more specifically, deep convolution neural networks (DCNN), has been broadly applied in vision tasks, such as detecting and tracking moving objects from remotely sensed images, surveillance videos, and autonomous robots \cite{chen2014vehicle,li2016deep,erhan2014scalable,lecun2015deep}. A great challenge in such tasks is the labor-intensive nature of preparing object-level labels, such as object class, which provides category information (image-level annotation), and object location – a bounding box (BBOX) showing the extent of each target object. This issue has drawn researchers’ attention to developing Weakly Supervised Objection Detection (WSOD) approaches \cite{wang2014weakly} that leverage weak labels (i.e., image-level annotation only) to achieve high-confidence object detection to alleviate the high cost associated with object labeling. 

Like strongly supervised OD networks, such as Faster RCNN \cite{ren2015faster}, a WSOD network typically consists of three key tasks in the object detection pipeline: (1) feature extraction: using a DCNN to extract low- to high-level features from the input images, (2) detection: relying on a region proposal network (RPN) to generate candidate region proposals containing the target objects, and (3) recognition: using a classifier to predict the object class. However, unlike the RPN used in Faster-RCNN, which was fed and trained with the ground-truth BBOX data, the challenge of WSOD is to predict region proposals that are highly likely to contain a target object without providing the BBOX information. 

There are two research directions in advancing WSOD: making improvement on the classifiers or developing new RPNs to generate more accurate proposals. Many existing works were reported to leverage image-level annotation to develop and refine proposal classifiers\cite{bilen2016weakly,cinbis2016weakly,deselaers2012weakly,diba2017weakly}. However, it is still very difficult for a WSOD network to achieve a level of prediction performance similar to that of strongly supervised approaches. One reason is that these works simply use the off-the-shelf region generation techniques such as selective search (SS) \cite{uijlings2013selective} or edge boxes (EB) \cite{zitnick2014edge}, resulting in limited performance increase. Recent studies \cite{hosang2015makes} have shown that the quality of proposals greatly affect the predictive performance of a network. Therefore, research taking the second direction – developing new RPNs has the potential to further boost the WSOD performance.

In this paper, we introduce a new solution in generating high-quality proposals by enabling a way of “learning from counting.” Unlike existing networks that need to generate a large number of candidate proposals and then select a subset from them, our proposed network can achieve better detection performance by automatically locating critical points on or near a target object. By generating a small number of proposals around the critical points, a set of high-quality proposals can be obtained and sent to the next WSOD stage. Our research is motivated by the use of a combined LSTM \cite{xingjian2015convolutional} and CTC in its outstanding capability in segmenting sequential data without per-frame labels, an idea similar to weakly supervised learning. To leverage this temporal classification network, we further apply a dimension reduction strategy to serialize input image into 1D sequential data and identify the critical object location leveraging count-based learning.  

To summarize, this work has made the following contributions: (1) It introduces for the first time the use of a Recurrent Neural Network (RNN)- LSTM as the proxy of a ‘weak’ RPN to improve WSOD performance. (2) The proposed RPN can easily be integrated into any WSOD network to generate high-quality proposals. (3) It enables a fully automated, end-to-end training framework with multiple independent data streams for region generation and classification to prevent the network from getting stuck in local optima. (4) Our method achieves state-of-the-art performance in WSOD.


\section{Literature Review}
\subsection{Weakly supervised object detection (WSOD)}
Existing efforts to improve the WSOD performance depends mainly on two research directions - developing better proposal classifiers and developing new RPNs to generate more accurate proposals. \cite{bilen2016weakly} developed an effective, end-to-end deep network for WSOD, in which a pre-trained CNN is used for feature extraction and two data streams are developed to undertake detection and recognition in parallel. However, this model tends to assign a higher score to a proposal that contains the most discriminative part of an object rather than the entire object. \cite{tang2017multiple} designed a strategy to assign the same image label for proposals that have significant overlaps with those receiving high scores during the weak supervision phase. An Online Instance Classifier Refinement (OICR) algorithm was then developed to use these proposals as pseudo ground-truths to classify the training images. Through continuous refinements, the proposed WSOD can achieve better instance recognition than the network in \cite{bilen2016weakly}. \cite{wan2019c} developed a C-MIL  (Continuation multiple instance learning) model to achieve WSOD using new loss functions. \cite{gao2018c} developed a Count-guided Weakly Supervised Localization (C-WSL) network to achieve high-confidence OD. This work addressed the issue of the tendency to draw a proposal containing multiple objects in existing weakly supervised detectors. \cite{gao2019c} leveraged segmentation maps with coupled multiple instance detection network (C-MIDN) to refine the proposals before sending them to the classifier, which also uses the OICR model. \cite{yang2019towards} integrated bounding-box regression (REG) into MIL as a single end-to-end network and enhanced the original feature map with implicit object location information by attention maps from images (guided attention module) (GAM). Two MIL approaches were adopted in this work: OICR and Proposal Cluster Learning (PCL)\cite{tang2018pcl}. 

All the above approaches aim to improve one or more stages of a WSOD network. However, their proposal generation processes mostly rely on mature techniques, such as SS or EB. However, \cite{hosang2015makes} and \cite{tang2018weakly} argue that the quality of proposals has a significant impact on the overall OD performance. Therefore, in recent years, more studies have been undertaken to improve proposal generation in RPNs. \cite{diba2017weakly} developed cascaded multiple networks with created class activation maps to infer better region proposals. \cite{tang2018weakly} proposed a two-stage network to improve the quality of generated proposals. The refined proposals are then sent to another WSOD \cite{tang2017multiple} to perform classification. Our research is also towards developing a RPN which can generate better proposals. But we take a very different approach - instead of relying on traditional object detection in a 2D realm, our approach converts the 2D object detection problem into a 1D sequence data segmentation problem and solves it by a novel use of LSTM and a count-based CTC. 

\subsection{Image labeling with LSTM and CTC} 
LSTM \cite{hochreiter1997long} is a type of RNN that was originally designed to model sequence data. LSTM can propagate information through lateral connections to model short- and long-term context dependencies. Regarding its usage in image processing tasks \cite{byeon2015scene,bell2016inside}, LSTM networks need to be extended from the temporal domain to the spatial domain and from single-dimensional learning to multi-dimensional learning \cite{graves2009offline}. \cite{visin2015renet} coupled four uni-dimensional RNNs to sweep across an image in four directions to replace the common convolutional-pooling layer. The network ensures that the output activation will appear in a specific location with respect to the whole image rather than a local context window of a CNN. In our work, we adopt the LSTM structure to identify critical points on a target object by taking advantage of its power in capturing global contexts and context dependencies in serialized data. 

CTC \cite{graves2006connectionist} is a type of scoring function and a neural network output designed for training RNNs to tackle sequence learning problems such as speech recognition. Instead of the need for per-frame labels, a CTC only needs “phoneme”-level labels whereby one phoneme can be mapped to multiple timeframes in the original speech audio. A CTC achieves this by transforming the network outputs into conditional probabilities and calculating the overall probabilities of all possible alignments that yield the same label sequence. It then finds the most probable sequence and corresponding alignment and uses that as the final output. The alignment can be represented by a set of segmented positions that separate each “phoneme” in a speech recognition problem. CTC’s ability to handle time sequence data and make predictions without the per-frame labels significantly broadened the applicability of RNNs.

This paper combines the LSTM and CTC to enable a novel WSOD by leveraging temporal classification. Next section introduces our methodology in detail.

\section{Methodology}

\begin{figure*}
\begin{center}
\fbox{\includegraphics[width=0.9\linewidth]{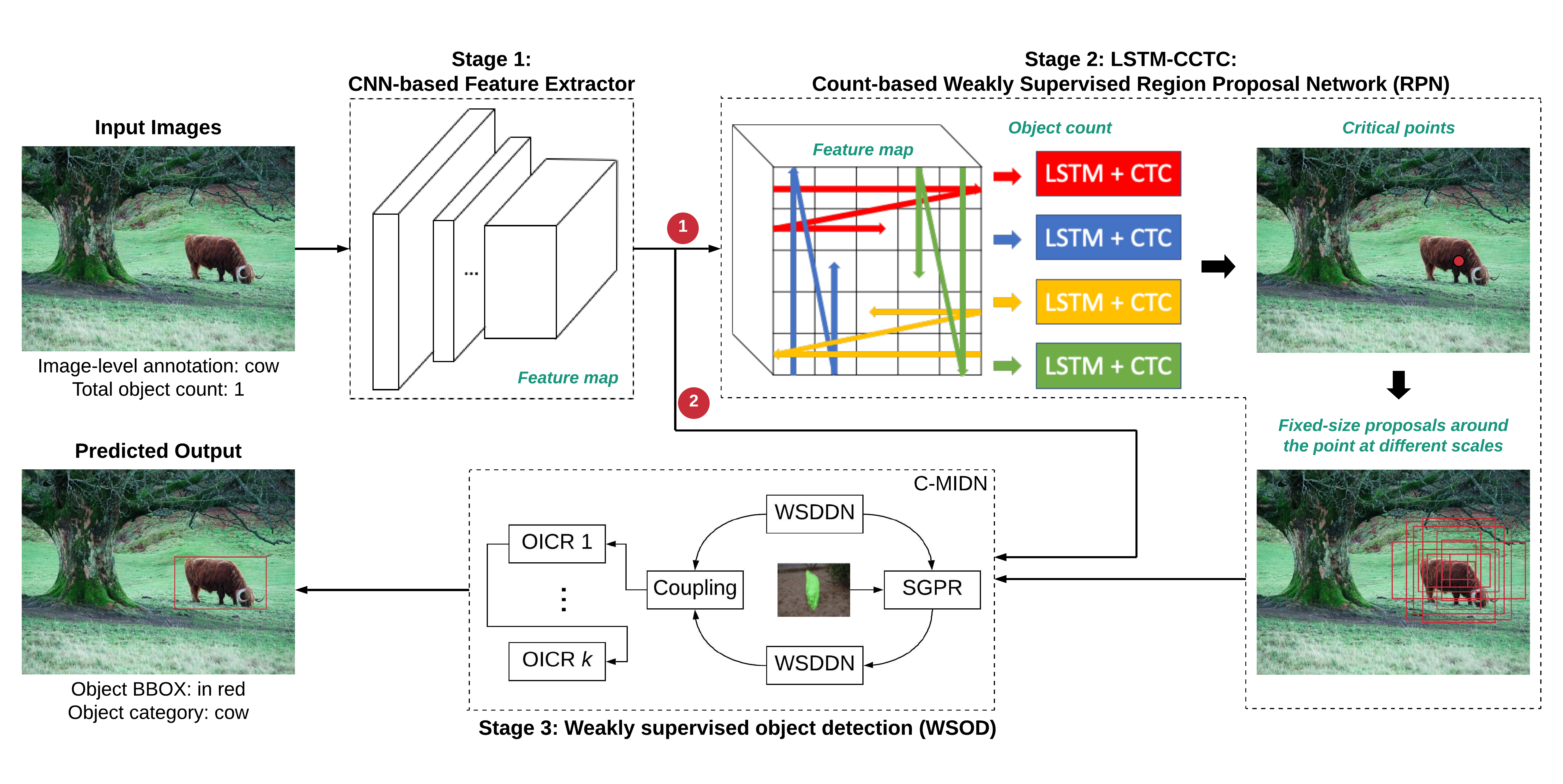}}
\end{center}
   \caption{Architecture of the proposed LSTM-CCTC for count-based WSOD.}
\label{fig1}
\end{figure*}

Figure \ref{fig1} illustrates the overall WSOD network with our proposed LSTM-CCTC as the RPN. The deep learning pipeline is divided into three main stages. First, a pre-trained CNN is used as the feature extractor. Second, we propose a new architecture for proposal generation. In this RPN, the resulting feature map will first be serialized into a 1D vector. This vector together with the total-class count will then be sent to the proposed LSTM-CCTC network to identify critical points falling on or near the target objects. A number of proposals with different aspect ratios and size will be generated around the critical points and sent to the last-stage of the WSOD network for proposal classification. While our proposed RPN (LSTM-CCTC) can be integrated into any WSOD, in this paper we used the one reported in \cite{gao2019c} for the classification task in Stage 3. We introduce details of each component in the following sections.

\subsection{Feature extractor}
Our feature extractor is built on a pretrained VGG16 neural network \cite{simonyan2014very} that is trained on ImageNet \cite{deng2009imagenet} using image-level labels. The feature map after the last convolutional block (convolution + ReLU) is fed into the proposed LSTM+CCTC for proposal generation and then the C-MIDN \cite{gao2019c} for classification. 


\subsection{Region proposal network (RPN)}
\begin{figure}
\begin{center}
\fbox{\includegraphics[width=0.7\linewidth]{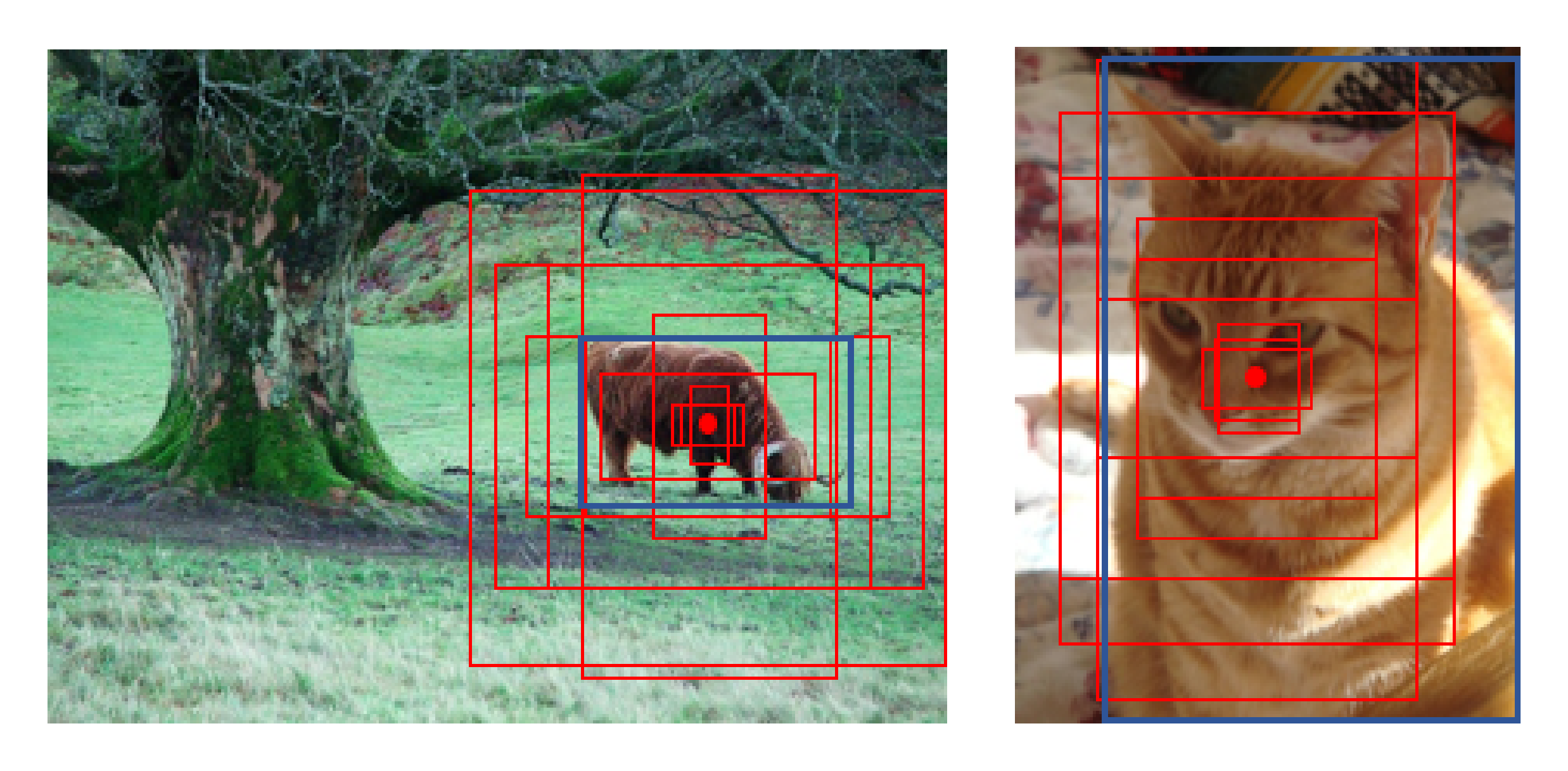}}
\end{center}
   \caption{Output from the proposed RPN (red point: segmented critical point by LSTM-CCTC, blue box: ground truth, red box: sample proposals.}
\label{fig2}
\end{figure}

Our PRN consists of four LSTMs that process the 1D feature map serialized by scan orders in four different directions (the four colored arrows on the feature map in Stage 2 of Figure \ref{fig1}). Each LSTM is connected to a CTC layer. The original CTC framework is designed for segmenting sequence data, while here we apply it to WSOD. To do so, we perform a transformation of the feature map with the size of \(n\times n\times k\) to a 1D vector with a length of \(n^2\). Each element is at \(1\times k\) in dimension. This conversion makes the feature map suitable to serve as the input of the LSTM + CTC network. Different from the traditional CTC model which trains on and predicts a sequence of (different) labels, our model is based on a count-based CTC, or CCTC, the goal of which is to inspect objects without the need to differentiate object type. So this LSTM-CCTC can also be considered as a binary segmentation problem. The object type is identified by the classifier at the next stage of the WSOD pipeline. After critical points are located, initial proposals with different aspect ratios (1:1, 2:1, 1:2, 1:3 and 3:1) are generated. Then, proposals are gradually enlarged while keeping at the same shape (Figure \ref{fig2}) until they hit the border of the feature map.

\subsection{Weakly supervised object detection (WSOD)}
In this stage, a region-based CNN is trained to classify proposals into different classes. In this work, we adopt C-MIDN \cite{gao2019c} as our WSOD network (Figure \ref{fig1}). This network couples two WSDDN \cite{bilen2016weakly} and contains iterative refinement in its instance classifier \cite{tang2017multiple}. The original WSDDN map proposal scores in a given image to an image-level classification confidence. Therefore, it can be trained solely under image-level supervision by optimizing a multi-class cross-entropy loss. However, the WSDDN often localizes the most discriminative object parts instead of the entire object because of the non-convex optimization. To address this issue, C-MIDN leverages a pair of WSDDNs, and the top-scoring proposals of the first WSDDN are removed from the input of the second WSDDN, preventing the second WSDDN from localizing the same proposal again. Furthermore, a segmentation map of the image is used to improve the robustness of the proposal removal process (Segmentation Guided Proposal Removal) (SGPR) to ensure that the full-context proposals will be kept. If the coverage rate between the segmentation map and the proposal is too small, it is more likely that there exist other tight proposals. Finally, proposals from the two WSDDNs are coupled to generate the final detection results. Next, the OICR training is adopted. Each stage is trained under the supervision of instance labels obtained from the previous stage. To obtain instance labels for supervision, given an image with a class label \(c\), the proposal \(j\) with the highest score for class \(c\) will be used as the pseudo ground-truth BBOX. Besides labeling \(j\), other proposals which have a high spatial overlap with \(j\) will also be labeled as class \(c\). The refinement strategy will give a preference to select the BBOX that contains the entire object instead of a part of it. It is very convenient to generate the pseudo ground-truth BBOX from our proposed RPN because these BBOXes (shown in Figure \ref{fig2}) share the same center, and many of them have a large overlap. 

\subsection{Implementation details}

In the feature extractor implementation, we replaced the last pooling layer with a SPP layer and combined it with the proposed RPN and a WSOD network (\cite{gao2019c}). In the RPN, the LSTMs are unidirectional with 2 layers, with the size at 512 for the input layer, and 256 for the hidden layer. The outputs are connected to the same fully-connected layer with the output size of 2 for binary object detection. We followed the original settings in \cite{gao2019c} for training and testing the WSOD network. During training, since the proposals from the first several iterations are noisy, we trained the proposed LSTM-CCTC for 20 epochs and then connected it with the WSOD network for an end-to-end training. The number of objects in each image is transferred into a sequence (e.g. 3 is transferred into 'ooo', where 'o' refers to a target object of any class) to train the LSTM-CCTC. All the initialization for newly added layers used Gaussian distributions with 0-mean and standard deviation 0.01. For the optimization, we used Stochastic Gradient Descent (SGD) with a min-batch size of 2. The learning rate is 0.001 for the first 200 epochs and 0.0001 for the following epochs. The momentum and weight decay are set to 0.9 and 0.0005 respectively.  

We also trained Fast R-CNN \cite{girshick2015fast} with top-scoring proposals generated by our proposed approach as the pseudo ground-truth. This is a common practice to improve the detection performance \cite{tang2017multiple,diba2017weakly,tang2018weakly,gao2019c}.

\section{Experiments}
In this section, we demonstrate the outstanding performance of our method through a series of experiments using benchmark datasets. We will also illustrate how different factors in the proposed method affect the prediction performance through ablation experiments.

\subsection{Experimental setup}
\hspace{0.5cm}\textbf{Datasets} We evaluated our method on PASCAL VOC 2007 and VOC 2012 datasets \cite{everingham2010pascal} which are two widely used benchmarks in WSOD. For both datasets, we combined training and validation images as the \emph{trainval} set for training and used test images for testing. We generated the object count for each image from the BBOX annotations to train our proposed RPN and used image-level labels to train the Stage 3 WSOD network. 

\textbf{Evaluation metrics} We used two performance metrics for evaluation: mean average precision (mAP) \cite{everingham2010pascal} and correct location (CorLoc) \cite{deselaers2012weakly}. mAP is a standard metric to evaluate the prediction accuracy and CorLoc measures the localization accuracy of a trained model. 

\textbf{Benchmarks} We compare our method with seven other popular WSOD networks, including (1) WSDDN \cite{bilen2016weakly}, one of the phenomenal CNN-based networks for WSOD; (2) OICR \cite{tang2017multiple}, an improved WSDDN with a classifier refinement algorithm OICR; (3) WSRPN \cite{tang2018weakly}, a CNN network focusing on generating high-quality proposals. This work shares the same goal as our proposed method; (4) C-WSL \cite{gao2018c}, which uses per-class count as the weak labels for WSOD; and three most recent state-of-the-art WSOD solutions: (5) C-MIDN \cite{gao2019c}, (6) C-MIL \cite{wan2019c}, and (7) MIL-OICR(PCL)+GAM+REG \cite{yang2019towards}. 

\subsection{Comparison with the state-of-the-arts}


\begin{table}
\tiny
\begin{center}

\begin{tabular}{l|l|c}
\thickhline
Method & aero\hspace{1pt} bike\hspace{2pt} bird\hspace{2pt} boat\hspace{0pt} bottle\hspace{2pt} bus\hspace{4pt} car\hspace{4pt} cat\hspace{2pt} chair\hspace{2pt} cow\hspace{2pt} table\hspace{2pt} dog\hspace{2pt} horse\hspace{0pt} mbike\hspace{0pt} person\hspace{0pt} plant\hspace{0pt} sheep\hspace{0pt} sofa\hspace{2pt} train\hspace{4pt} tv & mAP  \\
\hline
WSDDN \cite{bilen2016weakly} & 39.4\hspace{1pt} 50.1\hspace{1pt} 31.5\hspace{1.5pt} 16.3\hspace{2pt} 12.6\hspace{2pt} 64.5\hspace{1.5pt} 42.8\hspace{1pt} 42.6\hspace{1pt} 10.1\hspace{2pt} 35.7\hspace{2pt} 24.9\hspace{2pt} 38.2\hspace{2.5pt} 34.4\hspace{4pt} 55.6\hspace{4pt} 09.4\hspace{4pt} 14.7\hspace{2pt} 30.2\hspace{2pt} 40.7\hspace{2pt} 54.7\hspace{2pt} 46.9 & 34.8  \\

OICR \cite{tang2017multiple} & 58.0\hspace{1pt} 62.4\hspace{1pt} 31.1\hspace{1.5pt} 19.4\hspace{2pt} 13.0\hspace{2pt} 65.1\hspace{1.5pt} 62.2\hspace{1pt} 28.4\hspace{1pt} 24.8\hspace{2pt} 44.7\hspace{2pt} 30.6\hspace{2pt} 25.3\hspace{2.5pt} 37.8\hspace{4pt} 65.5\hspace{4pt} 15.7\hspace{4pt} 24.1\hspace{2pt} 41.7\hspace{2pt} 46.9\hspace{2pt} \textcolor{black}{64.3}\hspace{2pt} 62.6 & 41.2  \\

WSRPN \cite{tang2018weakly} & 57.9\hspace{1pt} 70.5\hspace{1pt} 37.8\hspace{1.5pt} 05.7\hspace{2pt} 21.0\hspace{2pt} 66.1\hspace{1.5pt} 69.2\hspace{1pt} 59.4\hspace{1pt} 03.4\hspace{2pt} 57.1\hspace{2pt} \textbf{57.3}\hspace{2pt} 35.2\hspace{2.5pt} 64.2\hspace{4pt} \textcolor{black}{68.6}\hspace{4pt} \textbf{32.8}\hspace{4pt} 28.6\hspace{2pt} 50.8\hspace{2pt} 49.5\hspace{2pt} 41.1\hspace{2pt} 30.0 & 45.3  \\

C-WSL+ODR \cite{gao2018c} & 62.7\hspace{1pt} 63.7\hspace{1pt} 40.0\hspace{1.5pt} 25.5\hspace{2pt} 17.7\hspace{2pt} \textcolor{black}{70.1}\hspace{1.5pt} 68.3\hspace{1pt} 38.9\hspace{1pt} 25.4\hspace{2pt} 54.5\hspace{2pt} 41.6\hspace{2pt} 29.9\hspace{2.5pt} 37.9\hspace{4pt} \textbf{64.2}\hspace{4pt} 11.3\hspace{4pt} 27.4\hspace{2pt} 49.3\hspace{2pt} 54.7\hspace{2pt} 61.4\hspace{2pt} \textbf{67.4} & 45.6  \\

C-WSL+ODR* \cite{gao2018c} & \textbf{62.9}\hspace{1pt} 64.8\hspace{1pt} 39.8\hspace{1.5pt} \textcolor{black}{28.1}\hspace{2pt} 16.4\hspace{2pt} 69.5\hspace{1.5pt} 68.2\hspace{1pt} 47.0\hspace{1pt} 27.9\hspace{2pt} 55.8\hspace{2pt} 43.7\hspace{2pt} 31.2\hspace{2.5pt} 43.8\hspace{4pt} 65.0\hspace{4pt} 10.9\hspace{4pt} 26.1\hspace{2pt} 52.7\hspace{2pt} 55.3\hspace{2pt} 60.2\hspace{2pt} 66.6 & 46.8  \\

C-MIL \cite{wan2019c} & \textcolor{black}{62.5}\hspace{1pt} 58.4\hspace{1pt} 49.5\hspace{1.5pt} \textcolor{black}{32.1}\hspace{2pt} 19.8\hspace{2pt} 70.5\hspace{1.5pt} 66.1\hspace{1pt} 63.4\hspace{1pt} \textcolor{black}{20.0}\hspace{2pt} 60.5\hspace{2pt} 52.9\hspace{2pt} 53.5\hspace{2.5pt} 57.4\hspace{4pt} 68.9\hspace{4pt} 08.4\hspace{4pt} 24.6\hspace{2pt} \textcolor{black}{51.8}\hspace{2pt} \textcolor{black}{58.7}\hspace{2pt} 66.7\hspace{2pt} 63.5 & 50.5  \\

MIL-OICR+GAM+REG \cite{yang2019towards} & \textcolor{black}{55.2}\hspace{1pt} 66.5\hspace{1pt} 40.1\hspace{1.5pt} \textcolor{black}{31.1}\hspace{2pt} 16.9\hspace{2pt} 69.8\hspace{1.5pt} 64.3\hspace{1pt} 67.8\hspace{1pt} \textcolor{black}{27.8}\hspace{2pt} 52.9\hspace{2pt} 47.0\hspace{2pt} 33.0\hspace{2.5pt} 60.8\hspace{4pt} 64.4\hspace{4pt} 13.8\hspace{4pt} 26.0\hspace{2pt} \textcolor{black}{44.0}\hspace{2pt} \textcolor{black}{55.7}\hspace{2pt} 68.9\hspace{2pt} 65.5 & 48.6  \\

MIL-PCL+GAM+REG \cite{yang2019towards} & \textcolor{black}{57.6}\hspace{1pt} 70.8\hspace{1pt} \textbf{50.7}\hspace{1.5pt} \textcolor{black}{28.3}\hspace{2pt} \textbf{27.2}\hspace{2pt} \textbf{72.5}\hspace{1.5pt} 69.1\hspace{1pt} 65.0\hspace{1pt} \textcolor{black}{26.9}\hspace{2pt} 64.5\hspace{2pt} 47.4\hspace{2pt} 47.7\hspace{2.5pt} 53.5\hspace{4pt} 66.9\hspace{4pt} 13.7\hspace{4pt} \textbf{29.3}\hspace{2pt} \textcolor{black}{56.0}\hspace{2pt} \textcolor{black}{54.9}\hspace{2pt} 63.4\hspace{2pt} 65.2 & 51.5  \\

C-MIDN \cite{gao2019c} & \textcolor{black}{53.3}\hspace{1pt} 71.5\hspace{1pt} 49.8\hspace{1.5pt} \textcolor{black}{26.1}\hspace{2pt} 20.3\hspace{2pt} 70.3\hspace{1.5pt} \textbf{69.9}\hspace{1pt} 68.3\hspace{1pt} \textbf{28.7}\hspace{2pt} \textbf{65.3}\hspace{2pt} 45.1\hspace{2pt} \textbf{64.6}\hspace{2.5pt} 58.0\hspace{4pt} 71.2\hspace{4pt} 20.0\hspace{4pt} 27.5\hspace{2pt} \textcolor{black}{54.9}\hspace{2pt} \textcolor{black}{54.9}\hspace{2pt} 69.4\hspace{2pt} 63.5 & 52.6  \\

LSTM-CCTC+OICR (Ours) & 56.9\hspace{1pt} \textbf{73.4}\hspace{1pt} \textcolor{black}{40.7}\hspace{1.5pt} 26.3\hspace{2pt} 16.4\hspace{2pt} 66.8\hspace{1.5pt} 65.3\hspace{1pt} \textcolor{black}{68.3}\hspace{1pt} 23.5\hspace{2pt} \textcolor{black}{58.3}\hspace{2pt} 40.9\hspace{2pt} 39.6\hspace{2.5pt} 40.4\hspace{4pt} 62.8\hspace{4pt} 24.6\hspace{4pt} 25.1\hspace{2pt} 48.3\hspace{2pt} 51.2\hspace{2pt} 57.7\hspace{2pt} 62.9 & \textcolor{black}{47.5}  \\

LSTM-CCTC+C-MIDN (Ours) & 60.2\hspace{1pt} \textcolor{black}{71.7}\hspace{1pt} \textcolor{black}{47.6}\hspace{1.5pt} \textbf{29.4}\hspace{2pt} 25.2\hspace{2pt} 72.1\hspace{1.5pt} 66.8\hspace{1pt} \textbf{69.6}\hspace{1pt} 27.4\hspace{2pt} \textcolor{black}{62.9}\hspace{2pt} 45.8\hspace{2pt} 58.3\hspace{2.5pt} 54.6\hspace{4pt} 67.8\hspace{4pt} 25.4\hspace{4pt} 26.8\hspace{2pt} \textbf{60.1}\hspace{2pt} \textbf{60.2}\hspace{2pt} 65.4\hspace{2pt} 64.2 & \textbf{53.1}  \\

\thickhline

OICR-Ens.+FRCNN \cite{tang2017multiple} & \textbf{65.5}\hspace{1pt}	67.2\hspace{1pt}	47.2\hspace{1.5pt}	21.6\hspace{2pt}	22.1\hspace{2pt}	68.0\hspace{1.5pt}	68.5\hspace{1pt}	35.9\hspace{1pt}	05.7\hspace{2pt}	63.1\hspace{2pt}	49.5\hspace{2pt}	30.3\hspace{2.5pt}	64.7\hspace{4pt}	66.1\hspace{4pt}	13.0\hspace{4pt}	25.6\hspace{2pt}	50.0\hspace{2pt}	57.1\hspace{2pt}	60.2\hspace{2pt}	59.0 & 47.0		\\

C-WSL+ODR+FRCNN \cite{gao2018c} & 61.9\hspace{1pt}	61.9\hspace{1pt}	48.6\hspace{1.5pt}	\textcolor{black}{28.7}\hspace{2pt}	23.3\hspace{2pt}	\textcolor{black}{71.1}\hspace{1.5pt}	71.3\hspace{1pt}	38.7\hspace{1pt}	\textbf{28.5}\hspace{2pt}	60.6\hspace{2pt}	45.4\hspace{2pt}	26.3\hspace{2.5pt}	49.7\hspace{4pt}	65.5\hspace{4pt}	07.2\hspace{4pt}	27.3\hspace{2pt}	54.7\hspace{2pt}	\textcolor{black}{61.6}\hspace{2pt}	\textcolor{black}{63.2}\hspace{2pt}	59.5 & 47.8 \\	

C-WSL+ODR*+FRCNN \cite{gao2018c} & 62.9\hspace{1pt}	68.3\hspace{1pt}	\textcolor{black}{52.9}\hspace{1.5pt}	25.8\hspace{2pt}	16.5\hspace{2pt}	\textcolor{black}{71.1}\hspace{1.5pt}	69.5\hspace{1pt}	48.2\hspace{1pt}	26.0\hspace{2pt}	58.6\hspace{2pt}	44.5\hspace{2pt}	28.2\hspace{2.5pt}	49.6\hspace{4pt}	66.4\hspace{4pt}	10.2\hspace{4pt}	26.4\hspace{2pt}	55.3\hspace{2pt}	59.9\hspace{2pt}	61.6\hspace{2pt}	\textcolor{black}{62.2} & 48.2 \\		

WSRPN-Ens.+FRCNN \cite{tang2018weakly} & 63.0\hspace{1pt}	69.7\hspace{1pt}	40.8\hspace{1.5pt}	11.6\hspace{2pt}	\textbf{27.7}\hspace{2pt}	70.5\hspace{1.5pt}	\textbf{74.1}\hspace{1pt}	58.5\hspace{1pt}	10.0\hspace{2pt}	66.7\hspace{2pt}	\textbf{60.6}\hspace{2pt}	34.7\hspace{2.5pt}	\textbf{75.7}\hspace{4pt}	\textbf{70.3}\hspace{4pt}	\textcolor{black}{25.7}\hspace{4pt}	26.5\hspace{2pt}	55.4\hspace{2pt}	56.4\hspace{2pt}	55.5\hspace{2pt}	54.9	& 50.4 \\	

C-MIL+FRCNN \cite{wan2019c} & 61.8\hspace{1pt}	60.9\hspace{1pt}	56.2\hspace{1.5pt}	28.9\hspace{2pt}	\textcolor{black}{18.9}\hspace{2pt}	68.2\hspace{1.5pt}	\textcolor{black}{69.6}\hspace{1pt}	71.4\hspace{1pt}	18.5\hspace{2pt}	64.3\hspace{2pt}	\textcolor{black}{57.2}\hspace{2pt}	66.9\hspace{2.5pt}	\textcolor{black}{65.9}\hspace{4pt}	\textcolor{black}{65.7}\hspace{4pt}	\textcolor{black}{13.8}\hspace{4pt}	22.9\hspace{2pt}	54.1\hspace{2pt}	\textbf{61.9}\hspace{2pt}	68.2\hspace{2pt}	\textbf{66.1}	& 53.1 \\

C-MIDN+FRCNN \cite{gao2019c} & 54.1\hspace{1pt}	\textbf{74.5}\hspace{1pt}	\textbf{56.9}\hspace{1.5pt}	26.4\hspace{2pt}	\textcolor{black}{22.2}\hspace{2pt}	68.7\hspace{1.5pt}	\textcolor{black}{68.9}\hspace{1pt}	\textbf{74.8}\hspace{1pt}	25.2\hspace{2pt}	64.8\hspace{2pt}	\textcolor{black}{46.4}\hspace{2pt}	\textbf{70.3}\hspace{2.5pt}	\textcolor{black}{66.3}\hspace{4pt}	\textcolor{black}{67.5}\hspace{4pt}	\textcolor{black}{21.6}\hspace{4pt}	24.4\hspace{2pt}	53.0\hspace{2pt}	59.7\hspace{2pt}	\textbf{68.7}\hspace{2pt}	58.9	& 53.6 \\

LSTM-CCTC+FRCNN (Ours) & 63.2\hspace{1pt}	\textcolor{black}{73.6}\hspace{1pt}	50.2\hspace{1.5pt}	\textbf{31.7}\hspace{2pt}	24.6\hspace{2pt}	\textbf{73.4}\hspace{1.5pt}	69.3\hspace{1pt}	\textcolor{black}{72.6}\hspace{1pt}	28.3\hspace{2pt}	\textbf{67.1}\hspace{2pt}	53.9\hspace{2pt}	\textcolor{black}{55.7}\hspace{2.5pt}	64.3\hspace{4pt}	66.1\hspace{4pt}	\textbf{26.8}\hspace{4pt}	\textbf{27.4}\hspace{2pt}	\textbf{61.2}\hspace{2pt}	60.8\hspace{2pt}	62.5\hspace{2pt}	59.4	& \textbf{54.6} \\

\thickhline
\end{tabular}
\end{center}
\caption{Result comparison in terms of AP (\%) and mAP (\%) on the PASCAL VOC 2007 \emph{test} set }
\label{voc2007map}
\end{table}

Table \ref{voc2007map} shows the AP and mAP evaluated on PASCAL VOC 2007 dataset. It can be seen that our method (LSTM-CCTC+C-MIDN) outperforms all other related works. In particular, our proposed model achieves better performance (7.8\% higher in mAP) than WSRPN \cite{tang2018weakly}, which also aims at improving the quality of generated proposals. Our proposed method also beats the count-guided C-WSL (C-WSL+ODR) \cite{gao2018c} by 7.5\% even with the use of weaker labels (total object count) than the per-class count used in C-WSL. We also integrated our proposed RPN into two well-performed WSOD frameworks: OICR\cite{tang2017multiple} and C-MIDN\cite{gao2019c}. The integrated networks yield a 6.3\% and a 0.6\% increase in mAP than \cite{tang2017multiple} and \cite{gao2019c}, respectively. This owes solely to the introduction of LSTM-CCTC for proposal generation.

Our model also achieves the state-of-the-art performance by training Fast R-CNN with pseudo ground-truths. The statistics on CorLoc are shown in Table \ref{voc2007cor}. The results show that our proposed WSOD achieves the best CorLoc among all methods compared, including those utilizing ensemble learning, such as WSRPN-Ens.+FRCNN\cite{tang2018weakly}.



\begin{table}
\tiny
\begin{center}

\begin{tabular}{l|l|c}
\thickhline
Method & aero\hspace{1pt} bike\hspace{2pt} bird\hspace{2pt} boat\hspace{0pt} bottle\hspace{2pt} bus\hspace{4pt} car\hspace{4pt} cat\hspace{2pt} chair\hspace{2pt} cow\hspace{2pt} table\hspace{2pt} dog\hspace{2pt} horse\hspace{0pt} mbike\hspace{0pt} person\hspace{0pt} plant\hspace{0pt} sheep\hspace{0pt} sofa\hspace{2pt} train\hspace{3pt} tv & mean\\

\hline

WSDDN \cite{bilen2016weakly} & 65.1\hspace{1pt} 58.8\hspace{1pt}	58.5\hspace{1.5pt}	33.1\hspace{2pt}	39.8\hspace{2pt}	68.3\hspace{1.5pt}	60.2\hspace{1pt}	59.6\hspace{1pt}	34.8\hspace{2pt}	64.5\hspace{2pt}	30.5\hspace{2pt}	43.0\hspace{2.5pt}	56.8\hspace{4pt}	82.4\hspace{4pt}	25.5\hspace{4pt}	41.6\hspace{2pt}	61.5\hspace{2pt}	55.9\hspace{2pt}	65.9\hspace{1pt}	63.7 & 53.5 \\

OICR \cite{tang2017multiple} & 81.7\hspace{1pt}	80.4\hspace{1pt}	48.7\hspace{1.5pt}	49.5\hspace{2pt}	32.8\hspace{2pt}	81.7\hspace{1.5pt}	85.4\hspace{1pt}	40.1\hspace{1pt}	40.6\hspace{2pt}	79.5\hspace{2pt}	35.7\hspace{2pt}	33.7\hspace{2.5pt}	60.5\hspace{4pt}	88.8\hspace{4pt}	21.8\hspace{4pt}	57.9\hspace{2pt}	76.3\hspace{2pt}	59.9\hspace{2pt}	75.3\hspace{1pt}	81.4 & 60.6		\\

C-WSL+ODR \cite{gao2018c} & \textbf{86.3}\hspace{1pt}	80.4\hspace{1pt}	58.3\hspace{1.5pt}	50.0\hspace{2pt}	36.6\hspace{2pt}	\textbf{85.8}\hspace{1.5pt}	86.2\hspace{1pt}	47.1\hspace{1pt}	42.7\hspace{2pt}	81.5\hspace{2pt}	42.2\hspace{2pt}	42.6\hspace{2.5pt}	50.7\hspace{4pt}	90.0\hspace{4pt}	14.3\hspace{4pt}	61.9\hspace{2pt}	\textcolor{black}{85.6}\hspace{2pt}	\textcolor{black}{64.2}\hspace{2pt}	\textcolor{black}{77.2}\hspace{1pt}	\textcolor{black}{82.4} & 63.3		\\

C-WSL+ODR* \cite{gao2018c} & 85.8\hspace{1pt}	81.2\hspace{1pt}	\textcolor{black}{64.9}\hspace{1.5pt}	50.5\hspace{2pt}	32.1\hspace{2pt}	84.3\hspace{1.5pt}	85.9\hspace{1pt}	54.7\hspace{1pt}	\textbf{43.4}\hspace{2pt}	80.1\hspace{2pt}	42.2\hspace{2pt}	42.6\hspace{2.5pt}	60.5\hspace{4pt}	90.4\hspace{4pt}	13.7\hspace{4pt}	57.5\hspace{2pt}	82.5\hspace{2pt}	61.8\hspace{2pt}	74.1\hspace{1pt}	\textcolor{black}{82.4}	& 63.5	\\

WSRPN \cite{tang2018weakly} & 77.5\hspace{1pt}	81.2\hspace{1pt}	55.3\hspace{1.5pt}	19.7\hspace{2pt}	\textcolor{black}{44.3}\hspace{2pt}	80.2\hspace{1.5pt}	\textbf{86.6}\hspace{1pt}	69.5\hspace{1pt}	10.1\hspace{2pt}	87.7\hspace{2pt}	\textbf{68.4}\hspace{2pt}	52.1\hspace{2.5pt}	\textbf{84.4}\hspace{4pt}	\textbf{91.6}\hspace{4pt} \textbf{57.4}\hspace{4pt}	\textbf{63.4}\hspace{2pt}	77.3\hspace{2pt}	58.1\hspace{2pt}	57.0\hspace{1pt}	53.8 & 63.8		\\

C-MIDN \cite{gao2019c} & \hspace{3pt}--\hspace{7pt}	--\hspace{7pt}	--\hspace{8pt}	--\hspace{9pt}	--\hspace{8pt}	--\hspace{7pt}	--\hspace{7pt}	--\hspace{8pt}	--\hspace{8pt}	--\hspace{8pt}	--\hspace{9pt}	--\hspace{8pt}	--\hspace{11pt}	--\hspace{10pt}	--\hspace{10pt}	--\hspace{8pt}	--\hspace{8pt}	--\hspace{9pt}	--\hspace{8pt}	-- & 65.0	\\

MIL-OICR+GAM+REG \cite{yang2019towards} & 81.7\hspace{1pt}	81.2\hspace{1pt}	58.9\hspace{1.5pt}	54.3\hspace{2pt}	\textcolor{black}{37.8}\hspace{2pt}	83.2\hspace{1.5pt}	\textcolor{black}{86.2}\hspace{1pt}	77.0\hspace{1pt}	42.1\hspace{2pt}	83.6\hspace{2pt}	\textcolor{black}{51.3}\hspace{2pt}	44.9\hspace{2.5pt}	\textcolor{black}{78.2}\hspace{4pt}	\textcolor{black}{90.8}\hspace{4pt} \textcolor{black}{20.5}\hspace{4pt}	\textcolor{black}{56.8}\hspace{2pt}	74.2\hspace{2pt}
\textbf{66.1}\hspace{2pt}	81.0\hspace{1pt}	\textbf{86.0} & 66.8		\\

MIL-PCL+GAM+REG \cite{yang2019towards} & 80.0\hspace{1pt}	83.9\hspace{1pt}	\textbf{74.2}\hspace{1.5pt}	53.2\hspace{2pt}	\textbf{48.5}\hspace{2pt}	82.7\hspace{1.5pt}	\textcolor{black}{86.2}\hspace{1pt}	69.5\hspace{1pt}	39.3\hspace{2pt}	82.9\hspace{2pt}	\textcolor{black}{53.6}\hspace{2pt}	61.4\hspace{2.5pt}	\textcolor{black}{72.4}\hspace{4pt}	\textcolor{black}{91.2}\hspace{4pt} \textcolor{black}{22.4}\hspace{4pt}	\textcolor{black}{57.5}\hspace{2pt}	83.5\hspace{2pt}	64.8\hspace{2pt}	75.7\hspace{1pt}	77.1 & 68.0		\\

C-MIDN \cite{gao2019c} & \hspace{3pt}--\hspace{7pt}	--\hspace{7pt}	--\hspace{8pt}	--\hspace{9pt}	--\hspace{8pt}	--\hspace{7pt}	--\hspace{7pt}	--\hspace{8pt}	--\hspace{8pt}	--\hspace{8pt}	--\hspace{9pt}	--\hspace{8pt}	--\hspace{11pt}	--\hspace{10pt}	--\hspace{10pt}	--\hspace{8pt}	--\hspace{8pt}	--\hspace{9pt}	--\hspace{8pt}	-- & 68.7	\\

LSTM-CCTC+OICR (Ours) & 76.9\hspace{1pt}	\textbf{85.7}\hspace{1pt}	60.0\hspace{1.5pt}	\textcolor{black}{52.5}\hspace{2pt}	32.1\hspace{2pt}	82.6\hspace{1.5pt}	84.2\hspace{1pt}	\textcolor{black}{80.5}\hspace{1pt}	37.7\hspace{2pt}	\textbf{88.1}\hspace{2pt}	45.5\hspace{2pt}	55.2\hspace{2.5pt}	56.4\hspace{4pt}	85.3\hspace{4pt}	37.1\hspace{4pt}	55.6\hspace{2pt}	78.3\hspace{2pt}	59.6\hspace{2pt}	69.4\hspace{1pt}	82.3 & \textcolor{black}{65.3}		\\

LSTM-CCTC+C-MIDN (Ours) & 78.5\hspace{1pt}	83.3\hspace{1pt}	63.9\hspace{1.5pt} \textbf{57.6}\hspace{2pt}	43.8\hspace{2pt}	85.5\hspace{1.5pt}	84.4\hspace{1pt}	\textbf{83.2}\hspace{1pt}	39.0\hspace{2pt}	87.8\hspace{2pt}	50.4\hspace{2pt}	\textbf{67.5}\hspace{2.5pt}	67.8\hspace{4pt}	90.2\hspace{4pt}	42.5\hspace{4pt}	46.3\hspace{2pt}	\textbf{87.9}\hspace{2pt}	62.5\hspace{2pt}	\textbf{84.4}\hspace{1pt}	84.6 & \textbf{70.0}		\\

\thickhline

OICR-Ens.+FRCNN \cite{tang2017multiple} & 85.8\hspace{1pt}	82.7\hspace{1pt}	62.8\hspace{1.5pt}	45.2\hspace{2pt}	43.5\hspace{2pt}	\textcolor{black}{84.8}\hspace{1.5pt}	87.0\hspace{1pt}	46.8\hspace{1pt}	15.7\hspace{2pt}	82.2\hspace{2pt}	51.0\hspace{2pt}	45.6\hspace{2.5pt}	83.7\hspace{4pt}	91.2\hspace{4pt}	22.2\hspace{4pt}	59.7\hspace{2pt}	75.3\hspace{2pt}	65.1\hspace{2pt}	76.8\hspace{1pt}	78.1 & 64.3		\\

C-WSL+ODR+FRCNN \cite{gao2018c} & 85.8\hspace{1pt}	78.0\hspace{1pt}	61.6\hspace{1.5pt}	\textcolor{black}{52.1}\hspace{2pt}	44.7\hspace{2pt}	81.7\hspace{1.5pt}	88.4\hspace{1pt}	49.1\hspace{1pt}	50.0\hspace{2pt}	82.9\hspace{2pt}	44.1\hspace{2pt}	44.4\hspace{2.5pt}	63.9\hspace{4pt}	92.4\hspace{4pt}	14.3\hspace{4pt}	60.4\hspace{2pt}	86.6\hspace{2pt}	68.3\hspace{2pt}	\textbf{80.6}\hspace{1pt}	82.8 & 65.6		\\

C-WSL+ODR*+FRCNN \cite{gao2018c} & \textbf{87.5}\hspace{1pt}	81.6\hspace{1pt}	\textcolor{black}{65.5}\hspace{1.5pt}	\textcolor{black}{52.1}\hspace{2pt}	37.4\hspace{2pt}	83.8\hspace{1.5pt}	87.9\hspace{1pt}	57.6\hspace{1pt}	\textcolor{black}{50.3}\hspace{2pt}	80.8\hspace{2pt}	44.9\hspace{2pt}	44.4\hspace{2.5pt}	65.6\hspace{4pt}	92.8\hspace{4pt}	14.9\hspace{4pt}	\textbf{61.2}\hspace{2pt}	83.5\hspace{2pt}	\textbf{68.5}\hspace{2pt}	77.6\hspace{1pt}	\textbf{83.5}	& 66.1	\\

WSRPN-Ens.+FRCNN \cite{tang2018weakly} & 83.8\hspace{1pt}	82.7\hspace{1pt}	60.7\hspace{1.5pt}	35.1\hspace{2pt}	\textbf{53.8}\hspace{2pt}	82.7\hspace{1.5pt}	\textbf{88.6}\hspace{1pt}	67.4\hspace{1pt}	22.0\hspace{2pt}	86.3\hspace{2pt}	\textbf{68.8}\hspace{2pt}	50.9\hspace{2.5pt}	\textbf{90.8}\hspace{4pt}	\textbf{93.6}\hspace{4pt}	\textcolor{black}{44.0}\hspace{4pt}	\textbf{61.2}\hspace{2pt}	82.5\hspace{2pt}	65.9\hspace{2pt}	71.1\hspace{1pt}	76.7 & 68.4		\\

C-MIDN+FRCNN \cite{gao2019c} & \hspace{3pt}--\hspace{7pt}	--\hspace{7pt}	--\hspace{8pt}	--\hspace{9pt}	--\hspace{8pt}	--\hspace{7pt}	--\hspace{7pt}	--\hspace{8pt}	--\hspace{8pt}	--\hspace{8pt}	--\hspace{9pt}	--\hspace{8pt}	--\hspace{11pt}	--\hspace{10pt}	--\hspace{10pt}	--\hspace{8pt}	--\hspace{8pt}	--\hspace{9pt}	--\hspace{8pt}	-- & 71.9	\\

LSTM-CCTC+FRCNN (Ours) & 84.5\hspace{1pt}	\textbf{84.7}\hspace{1pt}	\textbf{66.4}\hspace{1.5pt}	\textbf{59.1}\hspace{2pt}	43.3\hspace{2pt}	\textbf{86.8}\hspace{1.5pt}	\textcolor{black}{85.3}\hspace{1pt}	\textbf{88.5}\hspace{1pt}	\textbf{52.4}\hspace{2pt}	\textbf{87.0}\hspace{2pt}	60.3\hspace{2pt}	\textbf{71.4}\hspace{2.5pt}	80.3\hspace{4pt}	89.9\hspace{4pt}	\textbf{46.3}\hspace{4pt}	58.1\hspace{2pt}	\textbf{88.4}\hspace{2pt}	59.4\hspace{2pt}	78.8\hspace{1pt}	81.3 & \textbf{72.6}		\\

\thickhline
\end{tabular}
\end{center}
\caption{Result comparison in terms of CorLoc (\%) on the PASCAL VOC 2007 \emph{trainval} set }
\label{voc2007cor}
\end{table}


\begin{table}
\footnotesize
\begin{center}
  
  \begin{tabular}{l|cc}
    \thickhline
    Method & mAP & CorLoc \\
    \hline
    OICR \cite{tang2017multiple} & 37.9 & 62.1 \\
    WSRPN \cite{tang2018weakly} & 40.8 & 65.6 \\
    C-MIL \cite{wan2019c} & 46.7 & 67.4 \\
    MIL-OICR+GAM+REG \cite{yang2019towards} & 46.8 & 69.5 \\
    MIL-PCL+GAM+REG \cite{yang2019towards} & 45.6 & 68.7 \\
    C-MIDN \cite{gao2019c} & 50.2 & 71.2\\
    LSTM-CCTC+OICR (Ours) & 42.3 & 66.2 \\
    LSTM-CCTC+C-MIDN (Ours) & 50.5 & 72.5  \\
    \thickhline
    OICR-Ens.+FRCNN \cite{tang2017multiple} & 42.5 & 65.6 \\
    WSRPN-Ens.+FRCNN \cite{tang2018weakly} & 45.7 & 69.3 \\
    C-MIDN+FRCNN \cite{gao2019c} & 50.3 & 73.3 \\
    LSTM-CCTC+FRCNN (Ours) & 51.8 & 75.1 \\
    \thickhline
    
  \end{tabular}
  \end{center}
  \caption{Result comparison on the PASCAL VOC 2012 \emph{test} set}
  \label{voc12test}
\end{table}

The same experiments were conducted on VOC 2012 and results are shown in Table \ref{voc12test}. The results verify that our proposed model achieves better performance than the other popular WSOD models. 

\subsection{Ablation Study}
We also conducted ablation experiments on PASCAL VOC 2007 to analyze the impact of different factors on the performance of our proposed network. 

\textbf{Ways of feature map serialization} The first factor is ways of feature-map serialization, or more specifically, the number of scan orders used in serializing feature map at Stage 2. As described in Section 3.2, we made a point that applying LSTM to serialized data derived from multiple scanning directions will lead to identification of more (and accurate) critical points falling on or near a target object. This is based on the assumption that the “temporal” and contextual patterns exerted by different objects may be more predominantly shown in data serialized by different scan orders instead of a fixed one. We conducted experiments to apply only one way of serialization (direction in red in Figure \ref{fig1}), and two ways of serialization (directions in both red and black in Figure \ref{fig1}) and all four ways of serialization (Figure \ref{fig1}). The mAP and CorLoc we obtained for the three scenarios are 38.1\% mAP and 52.4\% CorLoc for scenario 1, 44.6\% mAP and 63.1\% CorLoc for scenario 2 and 53.1\% mAP and 70.0\% CorLoc for scenario 3. This result verifies our assumption. 

\textbf{Different proposal generation methods} Previous studies \cite{hosang2015makes} argued that the quality of proposals plays an important role in affecting the OD performance. This statement is especially true in a WSOD context when the exact BBOX labels are not available. Here, we compare our region generation method with typical methods such as SS and EB. To make a fair comparison, we integrated these comparing approaches into our learning framework (LSTM-CCTC+C-MIDN), with the replacement of Stage 2 by SS and EB, respectively. To ensure other networks to achieve their best possible performance within a reasonable training time, 2k proposals were generated for each method. Only an average of 200 proposals were generated by our proposed method instead. The results are: 52.6\% mAP and 68.7\% CorLoc for SS and 49.5\% mAP and 66.4\% CorLoc by EB. Our method (53.1\% mAP and 70.0\% CorLoc) clearly outperforms commonly used proposal generation techniques.  

\begin{figure}
\begin{center}
\fbox{\includegraphics[width=0.5\linewidth]{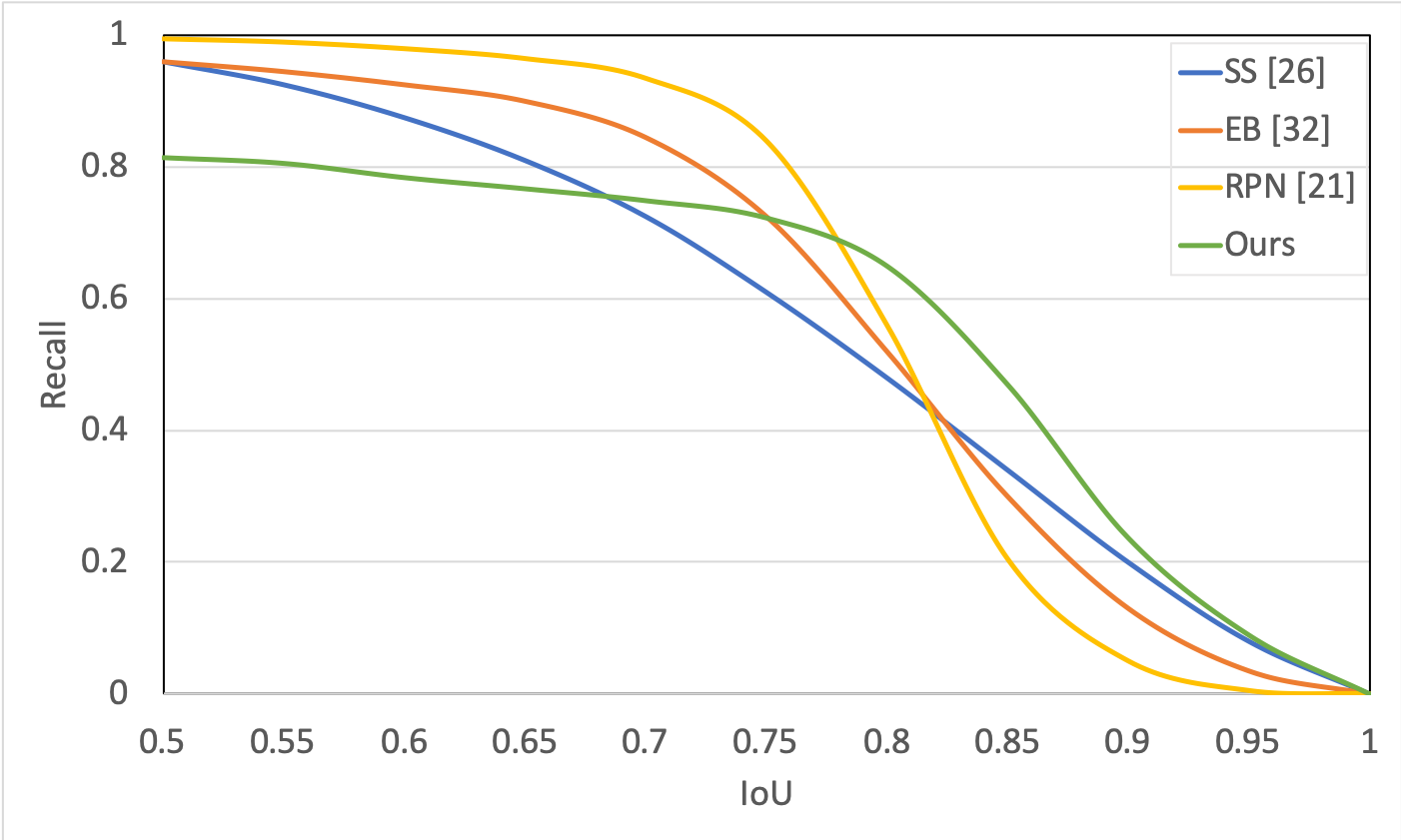}}
\end{center}
   \caption{Recall vs. IoU for different proposal methods on the VOC 2007 test set.}
\label{fig3}
\end{figure}

\textbf{Proposal recall} Figure \ref{fig3} shows the proposal recall with ground truth bounding boxes at different IoU levels \cite{tang2018weakly}. According to \cite{hosang2015makes, chavali2016object}, a high recall is not a necessary condition of high detection mAP. We simply use this result to measure the quantity and quality of the generated proposals by our proposed network. In figure \ref{fig3}, we observe that our recall is not as high as other methods when the IoU level is low. It is because our algorithm highly relies on finding objects in one-time scanning instead of an exhaustive search. Compared to an exhaustive search, our network might not be able to locate as many objects as they do, however, once an object is located, our network generates a proposal of much better quality. This can be seen in the figure that our curve decays slower than other methods when IoU level gets higher. Our algorithm proposes multiple candidate boxes with different sizes and ratios around an object location such that it is more likely to have a better-quality proposal. The recall curve of RPN \cite{ren2015faster} is used as a reference here showing the difference between strong supervision (with bounding box information) and weak supervision. 

This result proves that our network can correctly label the position of objects and once we have correct aspect ratios of objects, the network generates tight proposals pretty well. One concern of increasing the number of proposals is that it also increases the computation effort. However, our total number of proposals is still far less than traditional methods like SS \cite{uijlings2013selective} and EB \cite{zitnick2014edge}. Only an average of 200 proposals per image were generated by our proposed method. In addition, unlike other works which generate a fixed number of proposals for each image, our total number of proposals for each image is proportional to the number of interested objects in the image. Our network achieves a good trade-off between computation effort and detection accuracy.

\section{Conclusion and future work}
This paper introduces a new solution in developing a proposal generation network LSTM-CCTC to achieve high-confidence WSOD. We converted the 2D object recognition problem into a 1D sequential data segmentation problem by leveraging the power of a combined LSTM and CTC network. Specifically, we take advantage of LSTM in its ability of capturing temporal patterns and context dependencies in sequence data, and the CTC in segmenting sequence data without the need of providing frame-wise labels. An improvement is made upon the LSTM-CTC network to create a count-based CTC (CCTC) which will enable weak supervision through a total object count. Multiple data serialization methods are introduced to help more accurately identifying the segmented locations –critical points falling on or near a target object in the original image. Experimental results show that our proposed region generation method has achieved the state-of-the-art performance for WSOD. In the future, we will explore ways to further improve the location accuracy of the identified critical points in the LSTM-CCTC network for better proposal generation. 
\bibliography{bmvc_final}
\end{document}